\documentclass[10pt,twocolumn,letterpaper]{article}
\usepackage{cvpr}              %

\usepackage{graphicx}
\usepackage{amsmath}
\usepackage{amssymb}
\usepackage{booktabs}
\usepackage{xcolor}
\usepackage{array,makecell}
\usepackage{caption}
\usepackage{graphicx}
\usepackage{wrapfig,lipsum}
\usepackage{enumitem}
\usepackage{amsmath}
\usepackage{bm}
\usepackage{arydshln}
\usepackage[accsupp]{axessibility}
\usepackage{multirow}

\usepackage{algorithm}
\usepackage{algpseudocode}

\definecolor{cvprblue}{rgb}{0.21,0.49,0.74}
\usepackage[pagebackref,breaklinks,colorlinks,citecolor=cvprblue]{hyperref}

\def\paperID{6} %
\def\confName{CVPR}
\def\confYear{2024}

\newcommand\blfootnote[1]{%
  \begingroup
  \renewcommand\thefootnote{}\footnote{#1}%
  \addtocounter{footnote}{-1}%
  \endgroup
}

\begin{document}

\title{Adaptive Memory Replay for Continual Learning}

\author{
\textbf{
James Seale Smith\textsuperscript{*\,\,1,2}
\quad Lazar Valkov\textsuperscript{1}
\quad Shaunak Halbe\textsuperscript{2}
\quad Vyshnavi Gutta\textsuperscript{2}
}
\\
\textbf{
Rogerio Feris\textsuperscript{1}
\quad Zsolt Kira\textsuperscript{2}
\quad Leonid Karlinsky\textsuperscript{1}
}
\\
\normalsize
\textsuperscript{1}MIT-IBM Watson AI Lab 
\quad   \textsuperscript{2}Georgia Institute of Technology
}

\twocolumn[{%
\renewcommand\twocolumn[1][]{#1}%
\maketitle
\begin{center}
    \centering
    \includegraphics[page=1,width=\textwidth,trim={0 4.5cm 0 1.5cm},clip]{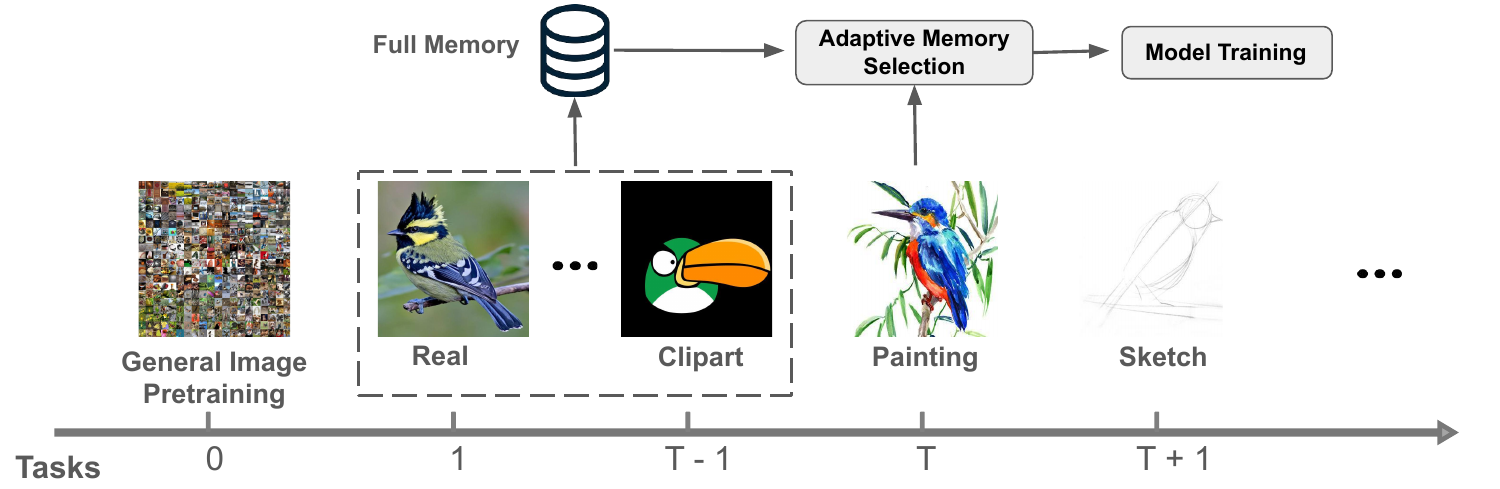}
    \captionof{figure}{\textbf{Adaptive memory replay for continual pre-training.} In our setting, we begin with general image pretraining (Task 0) and transition to learn different tasks (e.g., Real, Clipart, Painting, Sketch) with full memory access to all past task data. We choose relevant samples for model training, thereby minimizing catastrophic forgetting while efficiently updating the model with new data. Further, we \textit{replace} current task data with selected past task data, hence not adding training cost. 
    }
    \label{fig:key-idea}
\end{center}
}]

\begin{abstract}
Foundation Models (FMs) have become the hallmark of modern AI, however, these models are trained on massive data, leading to financially expensive training. Updating FMs as new data becomes available is important, however, can lead to `catastrophic forgetting', where models underperform on tasks related to data sub-populations observed too long ago. \blfootnote{*Work done during internship at MIT-IBM Watson AI Lab.}
This \textbf{continual learning} (CL) phenomenon has been extensively studied, but primarily in a setting where only a small amount of past data can be stored. We advocate for the paradigm where memory is abundant, allowing us to keep all previous data, but computational resources are limited. In this setting, traditional replay-based CL approaches are outperformed by a simple baseline which replays past data selected uniformly at random \cite{prabhu2023computationally}, indicating that this setting necessitates a new approach. We address this by introducing a framework of \textbf{adaptive memory replay for continual learning}, where sampling of past data is phrased as a multi-armed bandit problem. We utilize Bolzmann sampling to derive a method which dynamically selects past data for training conditioned on the current task, assuming full data access and emphasizing training efficiency.
Through extensive evaluations on both vision and language pre-training tasks, we demonstrate the effectiveness of our approach, which maintains high performance while reducing forgetting by up to $10\%$ at no training efficiency cost.
\end{abstract}

\section{Introduction}
\label{sec:intro}

The concept of the Foundation Models (FMs) \cite{bommasani2022opportunities} has recently gained popularity
and became ubiquitous in many downstream applications, including language \cite{touvron2023llama,brown2020language,openai2023gpt4}, vision \cite{radford2021learning,rombach2022highresolution,wang2022omnivlone}, and other application domains - advocating towards the `train-once-and-use-everywhere' paradigm shift in AI/ML. One of the most attractive features of FMs is their ability for Zero-Shot \cite{brown2020language} prompting, few-shot In-Context Learning (ICL) \cite{brown2020language,wei2023chainofthought}, and great transferability to any task \cite{openai2023gpt4,wang2022omnivlone}. This is due to their massive scale pre-training, often on billions \cite{schuhmann2022laion5b} or trillions \cite{openai2023gpt4,touvron2023llama} of data points. However, such power comes with high training costs. The pre-training data size is so large that normally each sample is observed only a few times or, as in language training, once (single epoch). Moreover, a common requirement for FMs is to rather frequently undergo `Extended Pre-Training' (EPT)
- a process of updating the model with new (massive) additional data intended to improve the model's temporal currency. During EPT the original pre-training data cannot be naively replayed as, given the massive size of both pre-training and EPT data, it  would effectively \textit{double} the EPT cost (naturally assuming a single epoch and 50\% replay mix for EPT as is customary in such cases). This would be prohibitive both in terms of the high cost (millions of dollars) as well as non-negligible negative environmental impact (extra heat emission).

However, neglecting past data during EPT is prone to the issue of \textit{catastrophic forgetting}~\cite{hsu2018re}, where models updated with new data tend to underperform on previously seen data. %
This leads to an important question: \emph{how can we adapt large-scale FM models to an ever-evolving world without compromising on performance or efficiency?}

\begin{figure}[t]
    \centering
    \includegraphics[page=1,width=0.48\textwidth,trim={10.5cm 4.5cm 10cm 3.9cm},clip]{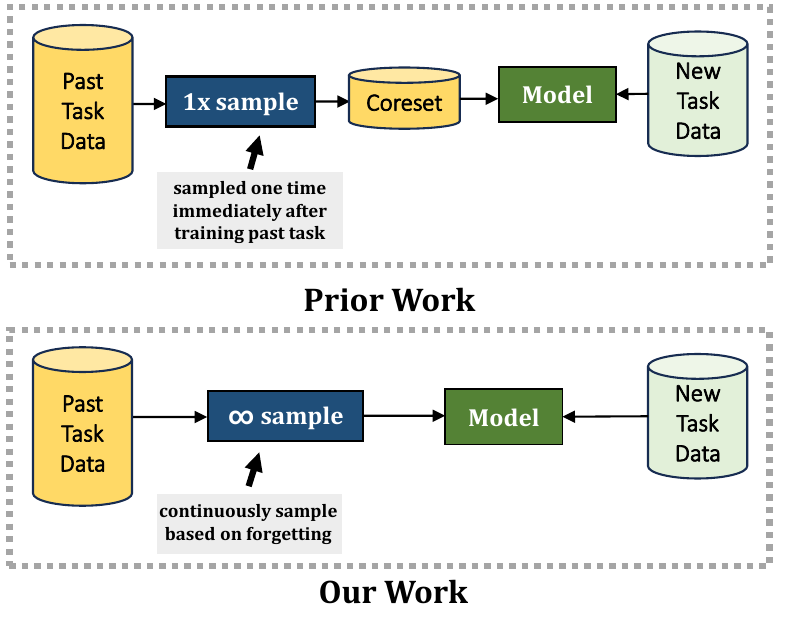}
    \caption{\textbf{Key difference of our work and prior work.} Prior work assumes that memory is expensive and constrains replay data to a fixed budget. Our work assumes memory is cheap and stores all replay data in memory, focusing on how to \emph{dynamically select the most useful replay data} for computation-budgeted replay.
    }
    \label{fig:key-idea-b}
\end{figure}
The realm of \textit{continual learning} offers some insights, but also limitations. While current benchmarks effectively highlight the challenge of catastrophic forgetting by training on non-overlapping data tasks sequentially, they are less applicable to (massive scale) EPT, as they either restrict themselves to limited memory storage (while in practical EPT all data, past and current, is usually available) and do not take into account the \textit{training cost of replay}. For practical EPT, we argue the cost impact needs to be \textit{minimal} in the sense that the `continual' EPT needs to have similar cost as `naive' EPT (disregarding old data and catastrophic forgetting issue). This is intuitive, as even the tiny overhead fraction due to replay will be applied as a factor to the training cost (measured in millions of dollars for the large-scale models \cite{openai2023gpt4}).

This new setting of restricted computation and unlimited storage of past data has recently been explored for continual learning of image classification tasks \cite{prabhu2023computationally}. The results of that work suggest that a simple baseline which randomly selects from all past data outperforms other replay-based CL methods. This creates a demand for a new approach which can better utilize all of the past data.

Taking inspiration from prior works~\cite{brignac2023improving,castro2018end,hayes2021selective} that have tried to select memory data intelligently by selecting the most representative samples of each memory replay dataset, we push the boundaries by considering not which past samples are the most representative (which is typically pre-decided before training future tasks), but rather \emph{which samples most effectively prevent forgetting conditioned on the current task data} (which is decided \emph{during the training of future tasks}).  This notion is based on the intuitive concept that the model retains full access to previously seen data and that the `optimal' replay data may be contingent upon the new data a model encounters during EPT.

We specifically propose an approach that dynamically adjusts the proportion of replay samples from each past task based on its propensity to be forgotten given the new task data. In such an \emph{adaptive memory replay for continual learning} (visualized in Figure~\ref{fig:key-idea}), our algorithm efficiently decides on the optimal allocation of memory replay samples among past tasks to minimize overall forgetting, under the vital consideration of how to do this without the requirement for drastic computation. We do this through a combination of bandit estimation and Boltzmann sampling from clusters of old datasets store in memory. We evaluate our replay strategy for both vision and language large-scale pre-training tasks. 
In particular, we propose and evaluate a \textit{zero-cost protocol} that includes intelligent selection of both data to replay and reduction in the new EPT data to compensate for the (relatively small) extra cost of the selection algorithm itself.

\textbf{In summary, we make the following contributions:}
\begin{enumerate}
\item We present an \emph{adaptive memory replay} for continual learning, a novel scheme inspired by a bandit estimation formulation that assumes full memory access and dynamically adjusts replay samples based on the new data, ensuring reduced forgetting.
\item Extensive evaluations demonstrate the efficacy of our method across both vision and language large-scale pre-training tasks.
\end{enumerate}

\section{Background and Related Work}
\label{sec:rl}

\noindent
\textbf{Continual Learning}: 
In the past few years, there has been significant progress in continual learning to alleviate catastrophic forgetting \cite{mccloskey1989catastrophic}. Regularization-based methods \cite{kirkpatrick2017overcoming,serra2018overcoming,li2017learning} modify the model parameters with additional regularization constraints to prevent catastrophic forgetting. They store no data but explore extra regularization terms in the loss function to consolidate previous knowledge. Rehearsal approaches \cite{rolnick2019experience,rolnick2019experience,buzzega2020dark} memorize or generate a small fraction of data points for previous tasks and utilizes them to retain the task knowledge. Importantly, what data to retain is decided during the task itself, and subsequently used throughout future tasks. Expansion approaches expand a model's architecture as new tasks are encountered; these are highly effective for applications where a model growing with tasks is practical~~\cite{ebrahimi2020adversarial,lee2020neural,lomonaco2017core50,maltoni2019continuous,Rusu:2016}. Our work does not consider these methods because the model parameters grow with the number of tasks, but acknowledge that the contributions could be incorporated into these approaches.

Recently, prompt-tuning methods such as \cite{wang2022dualprompt,wang2022learning,smith2023coda} outperformed rehearsal-based methods without using a replay
buffer by learning a small number of insertable model instructions or prompts.  Another line of research is the parameter isolation-based approaches \cite{mallya2018piggyback,yoon2017lifelong,houlsby2019parameter} which focus on freezing the task-specific parameters and growing new branches for new tasks. \cite{houlsby2019parameter} propose adapters which add a small number of parameters to the model for training on downstream tasks. Low-Rank Adaptation (LoRA) \cite{hu2021lora} extends on the above by using low-rank matrix counterparts of the original weights during fine-tuning, and keeps the actual weights frozen to further reduce inference costs.

\noindent
\textbf{Continual Learning in Transformers}:
The recent Vision Transformer (ViT) \cite{dosovitskiy2020transformers} has made a pure Transformer architecture scalable for large scale image classification and several works  \cite{li2022technical,douillard2022dytox,ermis2022memory} have successfully applied the Transformers architecture for continual learning. In \cite{li2022technical}, for each new task, the model is copied and fixed to be used as the teacher model in the distillation phase. In \cite{douillard2022dytox},  a unified model is learned by building upon a new architecture which dynamically expands the tokens processed by the last layer to mitigate forgetting. For each task, they learn a new task specific token per head using task-attention based decoder blocks. Recently,  \cite{ermis2022memory} proposes a method based on pre-trained Transformers while maintaining strict control of the memory usage and reaching state-of-the-art predictive performance. However the above methods either train a new transformer or need to fine-tune large pre-trained transformer models which requires significant compute in contrast to our objective of achieving optimal performance with limited compute.

\noindent
\textbf{Coreset Replay for Continual Learning}: 
Rehearsal-based methods use a memory buffer to store selective samples of the previous tasks. These samples are then replayed with new task data to prevent catastrophic forgetting. A notable rehearsal-based method, Experience Replay (ER) from memory \cite{rolnick2019experience} interleaves the previous task samples with the current task data for optimizing the network parameters. E2E~\cite{castro2018end} deploys a herding algorithm to bolster coreset representativeness of the past task training data. ERT \cite{buzzega2021rethinking} further extends ER by a balanced sampling strategy and bias control. Selective replay \cite{hayes2021selective} proposes task-based rehearsal strategies for sample selection based on class-margin boundary, minimum confidence etc. DER++ \cite{buzzega2020dark} mixes rehearsal with a distillation loss for preventing catastrophic forgetting. HAL \cite{chaudhry2021using} integrates ER with an additional objective of keeping the predictions on anchor points of past tasks intact. MIR \cite{rahaf2019online}, GDumb \cite{prabhu2020gdumb} and ASER \cite{shim2021online} store samples based on parameter updates, order of sample arrival and memory-based class boundaries respectively. Finally, ACE~\cite{brignac2023improving} explores various alternative population strategies to select coreset replay data. While the above methods consider the \emph{representativeness of data stored in the memory}, they fail to take into account the \emph{relationship between memory and the current task at hand}, as is explored in our work.

\section{Preliminaries} %
\label{sec:prelim}

\paragraph{Memory Replay}
In continual learning (CL)\footnote{In our setting, the model does \emph{not} have access to the task id during inference.}, the objective during task $T$, is to find parameters $\theta$ which minimize the loss $L$ over the current dataset $X_T$ and all previously seen datasets:

\begin{align} \label{eq:cl_objective}
O := \min_{\theta} \Bigg[ & \sum_{x \in X_T} \frac{L(x; \theta)}{|X_T|} +  \sum_{t=1}^{T-1} \sum_{x \in X_t} \frac{L(x; \theta)}{|X_t|} \Bigg]
\end{align}

Typically, CL approaches assume that past data cannot all be stored. Instead, experience replay approaches store a subset of past data from all previous tasks in a memory buffer $\mathcal{M} \subset \cup_{t=1}^{T-1} X_t$, where the size of $\mathcal{M}$ is much smaller compared to the combined number of data points from all past tasks. These approaches \cite{rolnick2019experience} use $\mathcal{M}$ to approximate the true objective (Eq. \ref{eq:cl_objective}) and minimize:
\begin{equation} \label{eq:typical_replay}
    \min_{\theta} \Bigg[
    \sum_{x \in X_T} \frac{L(x; \theta)}{|X_T|} + \alpha\sum_{x \in \mathcal{M}} \frac{L(x; \theta)}{|\mathcal{M}|}
    \Bigg]
\end{equation}

where $\alpha$ is a hyper-parameter. The memory buffer is updated after each task but the total number of stored items is constant.
The resulting method's computational requirements scale well with the number of learned tasks, but the limited size of $\mathcal{M}$ means that it becomes less effective at representing past data, as the number of learned tasks increases.

\paragraph{K-armed Bandits} The (stochastic) K-armed bandit problem \cite{bubeck2012regret} considers a setting in which there are $K$ available actions, referred to as \textit{arms}. Performing one of the actions returns a stochastic reward drawn from an unknown distribution. The problem is selecting a number of actions in a way which minimizes the expected regret, defined as the expected difference between the rewards obtained by always choosing the optimal action and the rewards obtained by following our strategy. At each step, a bandit strategy approximates the parameters of the reward distribution of each of the $K$ actions. Thereafter, the strategy needs to select an action to perform. One such strategy is \textit{Boltzmann Exploration} \cite{kaelbling1996reinforcement} which computes the mean of the observed rewards for each action, and then uses all means to define a categorical distribution, from which the choice of action is drawn. Finally, if the action's distributions change between steps, the bandit problem is referred to as \textit{non-stationary} \cite{zhao2020simple}.
\section{Adaptive Memory Replay}
\label{sec:method}
\begin{figure*}[t]
    \centering
    \includegraphics[page=1,width=\textwidth,trim={1.5cm 6.6cm 0.5cm 0.6cm},clip]{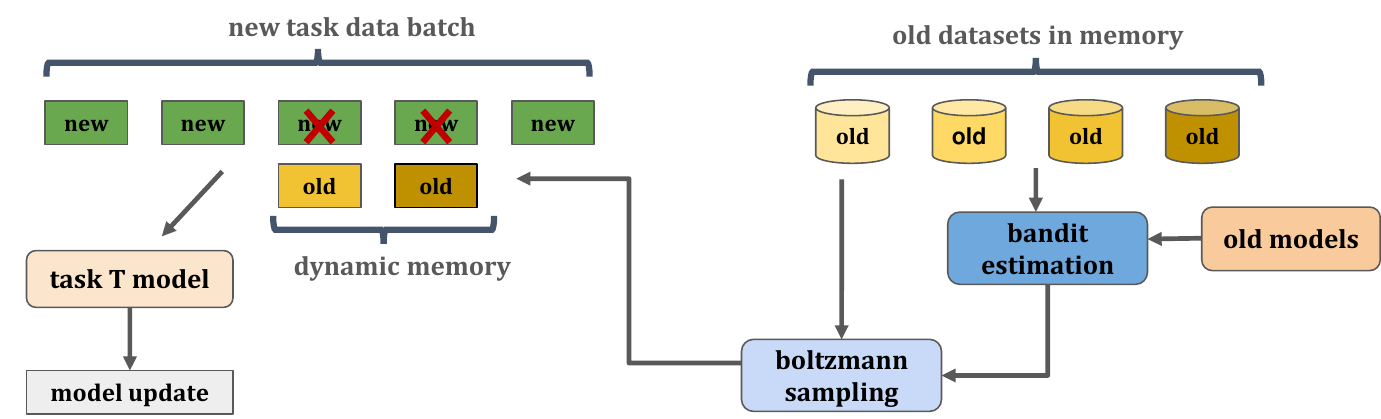}
    \caption{\textbf{Overview of our adaptive memory replay approach.} New task data is integrated with selectively rehearsed old data from full replay memory to update the task model. Unlike simple iid replay, our rehearsal data is chosen through a combination of bandit estimation and Boltzmann sampling from clusters of old datasets stored in memory. To reduce computation costs associated with data replay, we randomly discard samples from the training data to be replaced with the selected replay data. This ensures a cost-effective balance between incorporating new information and retaining knowledge of previous tasks, thus mitigating catastrophic forgetting with minimal computational overhead.
    }
    \label{fig:method}
\end{figure*}

In this section, we first modify the typical CL setting by challenging the restrictive assumption that past data cannot be accessed. Second, we modify the objective which we minimize, in order to better reflect the new CL setting. Third, we link the resulting problem to that of multi-armed bandit allocation \cite{bubeck2012regret} and detail our approach.

\subsection{Replay: A New Perspective}
\label{sec:method:new}

For FM extended pre-training, we challenge the common CL assumption that past data is unavailable and alter the common CL setting after making two observations. First, data storage is cheap, therefore it is possible to store and sample from any dataset we have seen before. Second, computation is expensive, meaning that we cannot re-train on all of the past data in memory. Following this insight, we modify the CL setting so that all of the past data can be stored, but a CL algorithm's computational demands to access and use the past data need to be constant and compensated during training to lead to zero extra cost (compared to naive training only on new data).

Being able to access all of the past data naturally leads to a rehearsal-based approach where one can only replay a limited subset $\mathcal{M}$, due to computational constraints. In contrast to the typical memory replay solutions, $\mathcal{M}$ is not fixed while training on a new task, but is instead allowed to adapt to the new data and the changing model while maintaining training efficiency. Therefore, we refer to it as an \textit{adaptive memory buffer}.

We begin developing our approach by modifying the objective function to better reflect our goal. First, as detailed in the Appendix, we express the main objective (Eq. \ref{eq:cl_objective}) in terms of its forgetting $\mathcal{F}$ on past data, compared to the performance of the optimal parameters for the previous task:
\begin{align}
O = \min_{\theta} \Bigg[ & \sum_{x \in X_T} \frac{L(x; \theta)}{|X_T|} \nonumber
 +  \sum_{t=1}^{T-1}\sum_{x \in X_t} \frac{\mathcal{F}(x; \theta)}{|X_t|} + C \Bigg]
\end{align}

where $C$ is a constant.
This change reflects the fact that we fine-tune the previously optimal model on the new task and that our focus is on minimizing forgetting, rather than improving our performance on past data. Next, we define the optimal memory buffer $\mathcal{M}^*(\theta)$ as the subset of past data for which our model has the maximum forgetting. We then following replay-based methods (Eq. \ref{eq:typical_replay}) and similarly optimize:
\begin{align} \label{eq:er_objective_mod}
\min_{\theta} \Bigg[ & \sum_{x \in X_T} \frac{L(x; \theta)}{|X_T|}
+ \alpha \sum_{x \in \mathcal{M}^*(\theta)} \frac{\mathcal{F}(x; \theta)}{|\mathcal{M}^*(\theta)|} + C \Bigg]
\end{align}

Minimizing this objective leads to our rehearsal-based CL algorithm with adaptive memory $M^*(\theta)$ which replays the past data points $x \in \mathcal{M}^*(\theta)$ that have currently been forgotten the most. Importantly, to minimize the overhead which replaying imposes on the training process, we keep the number of computed gradients constant by discarding a portion of the data on the new task. To do this, we remove $|M^*(\theta)|$ randomly selected data points for each batch of inputs from the new task, thus keeping the total number of processed inputs the same.

In order to compute Eq. $\ref{eq:er_objective_mod}$, we need to be able to select $\mathcal{M}^*(\theta)$ --- the subset of all past data with the highest forgetting. This subset changes as we update the parameters $\theta$, and it is computationally infeasible to evaluate the forgetting of each of the past data points. Instead, we seek to divide all past data into clusters $A_i$ of items expected to have similar forgetting values.

This allows us to infer the forgetting values of the data points in a cluster, based on a small number of evaluations, and use this to select data points for our adaptive memory buffer which exhibits a high amount of forgetting. Currently, we place all of the data from the same previous task into the same cluster, i.e. $A_i = X_i$, and assume that it would exhibit similar forgetting values. We leave more elaborate clustering techniques for future work.

\subsection{Adaptive Memory as a Bandit Optimization}
\label{sec:method:bandit}

Formally, we divide all past data into $K$ disjoint subsets $A_i$, s.t. $\cup_{i=1}^K A_i = \cup_{i=1}^{T-1}X_i$, where the forgetting of an input is distributed according to a subset-specific distribution: $\mathcal{F}(x) \sim \mathcal{D}_i$, for $x \in A_i$. For the parameters $\theta_j$ at training iteration $j$, we would like to select a subset $\mathcal{M}(\theta_j)$ which exhibits close to the worst forgetting, minimizing the following quantity:
$$r(\mathcal{M}(\theta_j)) := \sum_{x \in \mathcal{M^*}(\theta_j)} \mathcal{F}(x; \theta_j) - \sum_{x \in \mathcal{M}(\theta_j)} \mathcal{F}(x; \theta_j).$$

We frame this as a non-stationary K-armed bandit (KAB) problem \cite{slivkins2019introduction}, where pulling an arm and receiving a reward corresponds to sampling a data point from a cluster and evaluating its forgetting. Then, at each training iteration, we have to choose which of the K arms to pull in order to select $|\mathcal{M}|$ data points with maximum forgetting. As we select different $M$ over all training steps, we would like to reduce the \textit{expected regret}: $\mathbb{E} R:=\mathop{\mathbb{E}} [\sum_j  r(\mathcal{M}(\theta_j))]$, which is the expected difference in forgetting values between $\mathcal{M}$ and $\mathcal{M}*$ over all training steps. In this work, we implement the Boltzmann Exploration \cite{kaelbling1996reinforcement} approach which, at training step $j$, approximates the mean reward of each arm, denoted by $\mu_i^{(j)}$, and then uses all arms' means as parameters for a categorical distribution over the choice of arms to pull. This distribution is then used to sample $|\mathcal{M}|$ arms and in turn sample the adaptive memory buffer $\mathcal{M}(\theta_j)$.

To approximate the mean forgetting values $\mu_i^{(j)}$ of cluster $i$ at training step $j$, we first sample a small number of data points from the cluster $A_i$ and evaluate the average of their forgetting values --- $\bar{\mathbf{f}}^{(j)}_i$. We would like to compute $\mu_i^{(j)}$ based on the previously computed mean value $\mu_i^{(j-1)}$ and the currently computed forgetting average $\bar{\mathbf{f}}^{(j)}_i$. However, we note that the forgetting values depend on our model's parameters $\theta_j$, thus change between training iterations. We account for this by using a moving average, which is used for KAB when the underlying distributions are non-stationary \cite{sutton2018reinforcement}:
$\mu_i^{(j)} = \beta\bar{\mathbf{f}}^{(j)}_i + (1-\beta)\mu_i^{(j-1)}$.

Once we have approximated the mean forgetting values for all clusters, we use them to create a categorical distribution over the choice of clusters, with the help of the tempered softmax function \cite{hinton2015distilling}. We compute: $p(A_i)^{(j)} = \exp\{ \mu_i^{(j)}/t\}/Z$, where $t$ is the temperature hyperparameter and $Z$ is the normalization constant. Afterwards, we use this distribution to sample $|\mathcal{M}|$ cluster indices. Finally, we sample one input from each selected cluster, uniformly at random, and combine the samples to create the adaptive memory buffer for the current training step. 
\textbf{Our full method is summarized in Algorithm~\ref{alg:adaptive_memory_replay}.}

\begin{algorithm}[t]
\caption{Adaptive Memory Replay}
\label{alg:adaptive_memory_replay}
\begin{algorithmic}[1]
\Require Pre-trained model with parameters $\theta$, datasets $\{X_i\}_{i=1}^{T}$, constant $C$, learning rate $\eta$, replay buffer size $|\mathcal{M}|$, and temperature $t$
\Ensure Updated model $\theta$ that has minimal forgetting
\State Initialize replay buffer $\mathcal{M} \gets \emptyset$
\State Initialize clusters $A_i \gets X_i$ for each past dataset
\State Initialize mean forgetting $\mu_i^{(0)} \gets 0$ for each cluster $A_i$
\For{each training iteration $j$}
    \State Sample a batch $B$ from new data $X_T$
    \For{each cluster $A_i$}
        \State Sample points from $A_i$ to est. forgetting $\bar{\mathbf{f}}^{(j)}_i$
        \State Update means $\mu_i^{(j)} \gets \beta\bar{\mathbf{f}}^{(j)}_i + (1-\beta)\mu_i^{(j-1)}$
    \EndFor
    \State Compute distribution $p(A_i)^{(j)}$ using softmax
    \State Sample based on $p(A_i)^{(j)}$ to fill buffer $\mathcal{M}(\theta_j)$
    \State Remove $|\mathcal{M}(\theta_j)|$ random data points from $B$
    \State Update model to minimize objective fn (Eq.~\ref{eq:er_objective_mod})
\EndFor
\end{algorithmic}
\end{algorithm}

\section{Experiments}
\label{sec:exp}

\begin{table*}[t]
\caption{\textbf{Results on DomainNet~\cite{domainnet}}. Results are included for 6 tasks. Final Loss and Forgetting are reported using test data loss across all tasks after training on the entire sequence, normalized by the Oracle (full re-training on all datasets at each task). Training time for each approach is normalized using the Oracle training time. The `0-cost' result indicates the number of training steps of our approach is reduced to align with the training time of naive fine-tuning.}
\centering
\label{tab:vison:domainnet}
\begin{tabular}{c||c c||c c c} 
\hline 
\multirow{2}{*}{\thead{\\Approach}} & \multicolumn{2}{c||}{\thead{Normalized MAE Loss}} & \multicolumn{3}{c}{\thead{Normalized Training Time}} \\
\cline{2-6}
 & \thead{Final Loss\\($\downarrow$)} & \thead{Forgetting\\($\downarrow$)} & \thead{Total} & \thead{Selecting\\Replay Data} & \thead{Training\\Model} \\
\hline
Oracle & 0.0\% & 0.0\%  & 100.0\% & 0.0\% & 100.0\%   \\ 
Base & 100\% & 0.0\%  & 0.0\% & 0.0\% & 0.0\%   \\ 
\hline
Naive & 54.73\% & 70.95\% & 34.19\% & 0.0\% & 34.19\%   \\ 
Standard Rehearsal & 30.41\% & 22.97\% & 34.61\% & 0.0\%  & 34.61\%  \\ 
\hdashline
Our Rehearsal & \textbf{23.65\%} & \textbf{4.39\%} & 36.07\% & 2.45\% & 33.61\%   \\ 
Our Rehearsal (0 Cost) & \textbf{\textcolor{teal}{26.69\%}} & \textbf{\textcolor{teal}{12.84\%}} & 34.26\% & 2.31\% & 31.93\%   \\ 
\hline
\end{tabular}
\end{table*}

\begin{table*}[t]
\caption{\textbf{Results on Medical MNIST~\cite{medmnist}}. Results are included for 5 tasks. Final Loss and Forgetting are reported using test data loss across all tasks after training on the entire sequence, normalized by the Oracle (full re-training on all datasets at each task). Training time for each approach is normalized using the Oracle training time. The `0-cost' result indicates the number of training steps of our approach is reduced to align with the training time of naive fine-tuning.}
\centering
\label{tab:vison:medical}
\begin{tabular}{c||c c||c c c} 
\hline 
\multirow{2}{*}{\thead{\\Approach}} & \multicolumn{2}{c||}{\thead{Normalized MAE Loss}} & \multicolumn{3}{c}{\thead{Normalized Training Time}}  \\
\cline{2-6}
 & \thead{Final Loss\\($\downarrow$)} & \thead{Forgetting\\($\downarrow$)} & \thead{Total} & \thead{Selecting\\Replay Data} & \thead{Training\\Model} \\
\hline
Oracle & 0.0\% & 0.0\%  & 100.0\% & 0.0\% & 100.0\%   \\ 
Base & 100\% & 0.0\%  & 0.0\% & 0.0\% & 0.0\%   \\ 
\hline
\hline
Naive & 82.36\% & 98.80\% & 35.22 & 0.0\% & 35.22   \\ 
Standard Rehearsal & 12.57\% & 5.19\% & 36.05\% & 0.0\% & 36.05\%  \\ 
\hdashline
Our Rehearsal & \textbf{9.73\%} & \textbf{1.61\%} & 37.39\% & 1.69\% & 35.71\%   \\ 
Our Rehearsal (0 Cost) & \textbf{\textcolor{teal}{11.44\%}} & \textbf{\textcolor{teal}{1.98\%}} & 35.52\% & 1.60\% & 33.92\%   \\ 
\hline
\end{tabular}
\end{table*}

\begin{table*}[t]
\caption{\textbf{Results on Synthetic Visual Concepts~\cite{syvic}}. Results are included for 4 tasks. Final Loss and Forgetting are reported using test data loss across all tasks after training on the entire sequence, normalized by the Oracle (full re-training on all datasets at each task). Negative forgetting indicates forward transfer (which is only present in this unique dataset). Training time for each approach is normalized using the Oracle training time. The `0-cost' result indicates the number of training steps of our approach is reduced to align with the training time of naive fine-tuning.}

\centering
\label{tab:vison:syn}
\begin{tabular}{c||c c||c c c} 
\hline 
\multirow{2}{*}{\thead{\\Approach}} & \multicolumn{2}{c||}{\thead{Normalized MAE Loss}} & \multicolumn{3}{c}{\thead{Normalized Training Time}}  \\
\cline{2-6}
 & \thead{Final Loss\\($\downarrow$)} & \thead{Forgetting\\($\downarrow$)} & \thead{Total} & \thead{Selecting\\Replay Data} & \thead{Training\\Model} \\
\hline
Oracle & 0.0\% & 0.0\%  & 100.0\% & 0.0\% & 100.0\%   \\ 
Base & 100\% & 0.0\%  & 0.0\% & 0.0\% & 0.0\%   \\ 
\hline
\hline
Naive & 52.11\% & 27.93\% & 38.87\% & 0.0\% & 38.87\%   \\ 
Standard Rehearsal & 35.86\% & -6.61\% & 41.96\% & 0.0\% & 41.96\%   \\ 
\hdashline
Our Rehearsal  & \textbf{34.82\%} & \textbf{-7.78\%} & 46.40\% & 44.94\% & 1.46\%   \\  
Our Rehearsal (0 Cost) & \textbf{\textcolor{teal}{35.36\%}} & \textbf{\textcolor{teal}{-7.43\%}} & 37.31\% & 1.36\% & 35.94\%   \\  
\hline
\end{tabular}
\end{table*}
\begin{table*}[t]
\caption{\textbf{Results on 5-task Causal Language Modeling benchmark}. Final Loss and Forgetting are reported using test data loss across all tasks after training on the entire sequence, normalized by the Oracle (full re-training on all datasets at each task). Training time for each approach is normalized using the Oracle training time. The `0-cost' result indicates the number of training steps of our approach is reduced to align with the training time of naive fine-tuning.}
\centering
\label{tab:language:x}
\begin{tabular}{c||c c||c c c} 
\hline 
\multirow{2}{*}{\thead{\\Approach}} & \multicolumn{2}{c||}{\thead{Normalized MAE Loss}} & \multicolumn{3}{c}{\thead{Normalized Training Time}} \\
\cline{2-6}
 & \thead{Final Loss\\($\downarrow$)} & \thead{Forgetting\\($\downarrow$)} & \thead{Total} & \thead{Selecting\\Replay Data} & \thead{Training\\Model} \\
\hline
Oracle & 0.0\% & 0.0\% & 100.0\% & 0.0\% & 100.0\%  \\ 
Base & 100\% & 0.0\% & 0.0\% & 0.0\% & 0.0\%   \\ 
\hline
Naive & 137.54\% & 180.36\% & 36.05\% & 0.0\% & 36.05\%   \\ 
Standard Rehearsal & 44.11\% & 76.97\% & 40.32\% & 0.0\% & 40.32\%   \\ 
\hdashline
Our Rehearsal & \textbf{12.88\%} & \textbf{38.06\%} & 45.76\% & 40.33\% & 5.42\%   \\ 
Our Rehearsal (0 Cost) & \textbf{\textcolor{teal}{14.60\%}} & \textbf{\textcolor{teal}{39.75\%}} & 36.80\% & 4.04\% & 32.76\%   \\ 
\hline
\end{tabular}
\end{table*}

We evaluate the efficacy of our proposed adaptive memory replay for continual learning of FM pre-training (i.e., extended pre-training) in both the vision and language domains. We utilize two distinct pre-trained models as our backbones for these experiments: a Vision Masked Autoencoder (MAE)~\cite{mae} pre-trained on ImageNet-1K for vision-related tasks, and LLaMA~\cite{llama} with 7 billion parameters for language experiments. The evaluation metrics for our experiments are twofold: test data \textbf{Final Loss} and test data loss \textbf{Forgetting}. The Loss metrics are standard for FM training (especially in Language Modeling) as they are known to correlate with downstream use due to the massive pre-training volumes and de-facto seeing most of the samples only once. These metrics are normalized between 0\% and 100\%, where 0\% represents an offline upper bound with all data trained independently and identically distributed (iid), and 100\% corresponds to the performance of the pre-trained model without any fine-tuning.

Because our primary contribution lies in our novel perspective of full-memory replay, we compare our approach with full memory-access iid data replay as opposed to typical continual learning methods. We hypothesize that coreset selection replay methods are effectively upper-bounded by full iid replay, given their goal of identifying the most representative data for replay (rather than our perspective of identifying the most forgotten data for replay as a function of the current data). Consequently, we do not compare our method against sampled data replay in our main results tables as we store all data in memory.

Furthermore, our experiments are designed to demonstrate the advantages of our adaptive memory replay approach over traditional iid replay, especially in terms of computational efficiency and reduced forgetting. We consider gains of our approach to be orthogonal to the realms of non-replay regularization-based continual learning methods, and thus these comparisons are not the main focus of our results. Besides, from the perspective of computational efficiency, recent work has found such approaches to be impractical for computationally bounded continual learning~\cite{ghunaim2023real}. However, we do discuss the interaction of our approach with different continual learning strategies like regularization methods and knowledge distillation in our Appendix.

The hyperparameters for our experiments were meticulously chosen based on a series of small task experiments. We update our model on $10,000$ new data examples per task. In the interest of computational resources for the larger Llama model, we approximate the training of all the model parameters with LoRA finetuning~\cite{hu2021lora} in the language modeling experiments. In our experience, conclusions attained for LoRA finetuning reflect the same in full model training. We use a learning rate of $2e-5$ for full model fine-tuning and $2e-4$ for LoRA-based  fine-tuning. For our proposed adaptive memory replay bandit scheme, we found that a temperature of $t=0.1$ and forgetting mean update ratio of $\beta = 0.01$ performed best. We compose our replay batches for both iid replay and our adaptive memory replay with a 1:1 ratio of replay data to new task training data. We conducted evaluations on a hold-out test dataset comprising 500 samples per dataset. Additional training details can be found in our Appendix.

\subsection{Results for Vision SSL}
\label{section:exp_vision}
In Tables~\ref{tab:vison:domainnet},\ref{tab:vison:medical},\ref{tab:vison:syn}, we benchmark our proposed approach on 3 different continual pre-training sequences composed of vision datasets. Our goal was to demonstrate the robustness of our findings with a variety of unique and practical dataset sequences. The first dataset is the DomainNet~\cite{domainnet} dataset (Table~\ref{tab:vison:domainnet}), containing 6 different domains of common objects. The next is the Medical MNIST dataset~\cite{medmnist} (Table~\ref{tab:vison:medical}), from which we sampled 5 standardized biomedical image datasets containing the highest number of samples. Finally, we use 4 attribute splits from the Synthetic Visual Concepts (SyViC) dataset~\cite{syvic} (\ref{tab:vison:syn}).

Our results demonstrate the advantages of our adaptive memory replay method in the vision domain. Our approach consistently outperforms full memory iid replay (which serves as an upper bound for other replay-based continual learning methods that sample from a limited coreset), achieving lower final loss and forgetting rates. The slight increase in normalized training time is negligible compared to the performance gains, and furthermore, we show a 0-cost result where we reduce the number of training steps of our approach to align with the training time of naive fine-tuning, and show that even this result outperforms iid replay in all three benchmarks. 

We note that our strongest performance gains come from the DomainNet results. The gains for the medical data sequence and synthetic data pre-training are much more modest, yet remain pronounced. The synthetic data sequence is interesting in that forward transfer (i.e., negative forgetting) appears in all results - however, our method still has more forward transfer compared to iid replay. In practical terms, our results imply that vision systems equipped with our continual learning strategy would exhibit improved robustness over time, adapting to new data without significant loss of prior knowledge or computational costs. We re-iterate that there is much more room for improvement from our full-memory continual learning perspective - advanced strategies can close the gap between our method and the upper bound with fixed time costs by exploring interesting questions such as \emph{how to better cluster the data} and \emph{which new data is more or less favorable to discard}.

\subsection{Results for Causal Language Modeling}
\label{section:exp_language}

Our approach is further affirmed through our language experiments using the Llama model. In Table~\ref{tab:language:x}, we benchmark on a 5-dataset sequence using datasets from Huggingface~\cite{hug}. These datasets were chosen based on the significant variations in loss observed post fine-tuning, thus providing a rigorous test for our approach. The datasets encompass a broad range of language tasks, ensuring that our results are representative of diverse language modeling scenarios.
Further specifics about these datasets are available in our Appendix. %

The performance of our adaptive memory replay in our language experiments mirrors the success observed in the vision tasks. We see a significant reduction in both forgetting and final loss compared to the iid full-memory replay. Furthermore, the `0-cost' variant of our method is particularly noteworthy, as it manages to retain a high level of performance without additional computational expenditure compared to naive fine-tuning. This aspect is crucial for applications where computational resources are limited, especially fitting LLM extended pre-training where due to high data volumes and enormous model sizes even a tiny fraction of extra cost is intolerable.

\subsection{Additional Analysis}
\label{section:exp_ablations_analysis}

\begin{figure}[t]
    \centering
    \includegraphics[width=.48\textwidth,trim={1cm 0 1cm 0},clip]{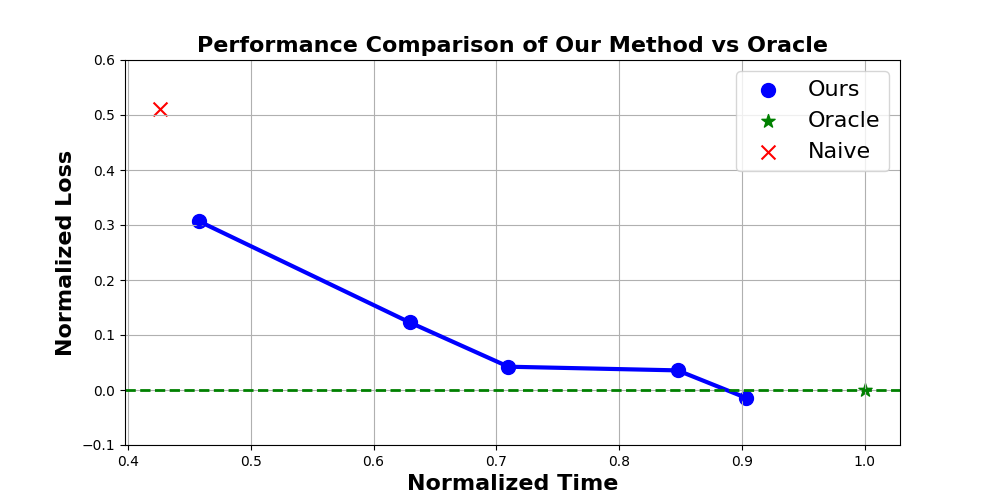}
     \caption{Final Loss vs Training Time for  adaptive memory replay vs Oracle using the Synthetic Visual Concepts~\cite{syvic} sequence.}
     \label{fig:analysis}
\end{figure}

In Figure~\ref{fig:analysis}, we present a comprehensive comparison of the final loss versus training time for our adaptive memory replay method against the Oracle, using the Synthetic Visual Concepts dataset sequence~\cite{syvic}. This plot demonstrates how our method converges towards the Oracle's performance as we increase the compute budget of our method (via using more replay samples and discarding fewer new task samples). With a limited budget, there is a notable difference in the final loss between our method and the Oracle. However, as training progresses, our method steadily approaches the Oracle's level of performance, matching and even outperforming Oracle (which as a fixed compute budget itself, pre-defined by the number of training steps we use) with a lower compute cost. The horizontal dotted line marks the point at which our approach reaches the Oracle's normalized loss, showcasing the efficiency of our method in terms of both loss minimization and computational time. This result is significant as it not only validates the effectiveness of our adaptive memory replay in reducing the final loss but also highlights its capability to achieve this with a substantially lower training time. 

\section{Conclusion}
\label{conclusion}

In this paper, we underscored the importance of adapting machine learning methodologies to the ever-evolving demands of real-world large-scale continual learning. Our findings, rooted in extensive evaluations across both vision and language tasks, validate the advantages of our \emph{adaptive memory replay for continual learning}. By dynamically selecting past training data samples, our method offers a nuanced balance, ensuring minimized forgetting without imposing costly computational requirements. Ultimately, we hope to inspire methodologies that are both computationally efficient and effective in real-world continual learning scenarios, where data access and computational resources are bound by practical constraints. This research trajectory serves as a stepping stone towards the development of continually learning systems that efficiently and intelligently utilize all available data, enhancing their learning and adaptability across a series of tasks throughout their life-cycle.
\paragraph{Limitations \& Future work.} Future work should focus on refining the adaptive memory replay mechanism, particularly exploring more sophisticated bandit-based selection strategies to further enhance the balance between retaining old knowledge and accommodating new information. The decision of which data to discard during the replay phase also warrants deeper investigation to avoid potential loss of critical information (and furthermore potential unrealized gains in discarding repetitive or similar samples). There is also a great need to develop advancing clustering techniques to capture the subtleties of data evolution which can lead to more representative memory buffers. In addition, bringing greater realism into continual learning models by incorporating real-world constraints and scenarios will be crucial, such as blurred task boundaries and online clustering. 

{\small
\bibliographystyle{ieeenat_fullname}
\bibliography{references}
}

\clearpage
\section*{Appendix}
\setcounter{figure}{0}
\setcounter{table}{0}
\renewcommand{\thetable}{\Alph{table}}
\renewcommand{\thefigure}{\Alph{figure}}
\renewcommand\thesection{\Alph{section}}
\appendix
\section{Method} \label{appendix-method}

This section shows how to express the CL objective (Eq.~1) in terms of the amount of forgetting. To start off, for task $T$, we denote the optimal parameters found on the previous task as $\theta_{T-1}^*$. Then, we define the forgetting for some parameter on some example to be positive if the loss on that example has increased: $\mathcal{F}(x; \theta) = \mathcal{L}(x; \theta) - \mathcal{L}(x; \theta^*_{T-1})$. Starting from our objective in Eq.~1, we write:

\begin{align*}
\min_{\theta} \Bigg[ & \sum_{x \in X_T} \frac{L(x; \theta)}{|X_T|} +  \sum_{t=1}^{T-1} \sum_{x \in X_t} \frac{L(x; \theta)}{|X_t|} \Bigg] \\
= \min_{\theta} \Bigg[ & \sum_{x \in X_T} \frac{L(x; \theta)}{|X_T|} 
 \\ &+  \sum_{t=1}^{T-1} \sum_{x \in X_t} \frac{L(x; \theta) - L(x; \theta^*_{T-1}) + L(x; \theta^*_{T-1})}{|X_t|} \Bigg] \nonumber \\
= \min_{\theta} \Bigg[ & \sum_{x \in X_T} \frac{L(x; \theta)}{|X_T|} 
 \\ &+  \sum_{t=1}^{T-1} \sum_{x \in X_t} \frac{L(x; \theta) - L(x; \theta^*_{T-1})}{|X_t|} \\ &+ \sum_{t=1}^{T-1} \sum_{x \in X_t} L(x; \theta^*_{T-1}) \Bigg] \nonumber \\
= \min_{\theta} \Bigg[ & \sum_{x \in X_T} \frac{L(x; \theta)}{|X_T|} \nonumber
 +  \sum_{t=1}^{T-1}\sum_{x \in X_t} \frac{\mathcal{F}(x; \theta)}{|X_t|} + C \Bigg] 
\end{align*}

Finally, we note that when minimizing the forgetting $\mathcal{F}(x; \theta) = \mathcal{L}(x; \theta) - \mathcal{L}(x; \theta^*_{T-1})$, only only needs to compute and minimize the loss on the new task $\mathcal{L}(x; \theta)$, since $\mathcal{L}(x; \theta^*_{T-1})$ is a fixed value. Therefore, we can optimize $\mathcal{F}$ without introducing extra computational demands to our training process.

\section{On Regularization Losses}
\label{appendix-other}

In our approach, we prioritize computational efficiency and focus on methods that do not incur additional computational costs. This decision is informed by the findings of Ghunaim \emph{et al.}~\cite{ghunaim2023real}, who demonstrate that both simple and advanced regularization-based continual learning techniques struggle to perform effectively under computational budget constraints. Moreover, their research suggests that simple experience replay is a more effective strategy in such scenarios. Thus, when extending such computational considerations to the setting of extended continual pre-training, we focus on \emph{outperforming iid experience replay without introducing any additional computational costs.} Furthermore, we consider gains of our approach to be orthogonal to the realms of non-replay regularization-based continual learning methods, and thus our method could potentially be integrated with these regularization techniques to enhance overall performance, offering a synergistic effect.

\section{Expanded Implementation Details}
\label{appendix-details}
We use A100 GPUs to generate all results. The hyperparameters for our experiments were meticulously chosen based on a series of small task experiments in which we use only used half of the number of tasks. We update our model on $10,000$ new data examples per task. In the interest of computational resources for the larger Llama model, we approximate the training of all the model parameters with LoRA finetuning~\cite{hu2021lora} in the language modeling experiments. In our experience, conclusions attained for LoRA finetuning reflect the same in full model training. We use a learning rate of $2e-5$ for full model fine-tuning and $2e-4$ for LoRA-based  fine-tuning. For LoRA-based fine-tuning, we use a rank of 8 for the Llama model experiments. For our proposed adaptive memory replay bandit scheme, we found that a temperature of $t=0.1$ and forgetting mean update ratio of $\beta = 0.01$ performed best. We compose our replay batches for both iid replay and our adaptive memory replay with a 1:1 ratio of replay data to new task training data. We conducted evaluations on a hold-out test dataset comprising 500 samples per dataset. We used a batch size of 128 and 16 for the Masked Autoencoder and Llama models, respectively, which was chosen based on GPU memory. For the Llama experiments, we leveraged low-precision training.

\section{Expanded Benchmark Details}
\label{appendix-benchmark}

In our main text, we evaluated the Masked Autoencoder model for three vision datasets. The first dataset is the DomainNet~\cite{domainnet} dataset, containing 6 different domains of common objects. The next is the Medical MNIST dataset~\cite{medmnist}, from which we sampled 5 standardized biomedical image datasets containing the highest number of samples. Finally, we use 4 attribute splits from the Synthetic Visual Concepts (SyViC) dataset~\cite{syvic}.

For the Llama model, we benchmarked on a 5-dataset sequence using datasets from Huggingface~\cite{hug}. The datasets involved in this sequence were \emph{banking77}~\cite{Casanueva2020}, \emph{wiki-cat-sum/animal}~\cite{perez-beltrachini-etal-2019-generating}, \emph{bigbio/hallmarks-of-cancer}~\cite{DBLP:journals/bioinformatics/BakerSGAHSK16}, \emph{big-patent}~\cite{DBLP:journals/corr/abs-1906-03741}, and \emph{wikitext}~\cite{merity2016pointer}.

\end{document}